\documentclass{article}

\usepackage{arxiv}

\usepackage{graphicx}
\usepackage{amsmath}
\usepackage{amssymb}
\usepackage{url}
\usepackage{booktabs}
\usepackage{cite}
\usepackage{xcolor}
\usepackage{caption}

\usepackage{algorithmicx}
\usepackage[utf8]{inputenc}
\usepackage{textgreek}
\usepackage{algpseudocode}
\usepackage{algcompatible}
\usepackage[ruled,vlined,linesnumbered]{algorithm2e}
\usepackage{amsmath}
\usepackage{tikz}
\usetikzlibrary{shapes.geometric, arrows.meta, positioning}

\tikzstyle{phase} = [rectangle, rounded corners,
                     text width=8cm, minimum height=1.2cm,
                     align=center, draw=black, fill=gray!10]
\tikzstyle{arrow} = [thick, ->, >=stealth]

\title{Lite VLA: Efficient Vision-Language-Action Control on CPU-Bound Edge Robots}

\author{
Justin Williams$^{1}$, Kishor Datta Gupta$^{2}$, Roy George$^{3}$, and Mrinmoy Sarkar$^{4}$\\[6pt]
$^{1,2,3}$Department of Cyber-Physical Systems, Clark Atlanta University, Atlanta, GA, USA\\
$^{4}$Siemens Corporation, Princeton, NJ, USA\\[6pt]
\texttt{$^{1}$justin.williams1@students.cau.edu}, 
\texttt{$^{2}$kgupta@cau.edu}, 
\texttt{$^{3}$rgeorge@cau.edu}, 
\texttt{$^{4}$mrinmoy.sarkar@siemens.com}
}

\renewcommand{\shorttitle}{\textit{arXiv} Template}

\begin{document}
\maketitle

\begin{abstract}
	The deployment of artificial intelligence models at the edge is increasingly critical for autonomous robots operating in GPS-denied environments where local, resource-efficient reasoning is essential. This work demonstrates the feasibility of deploying small Vision-Language Models (VLMs) on mobile robots to achieve real-time scene understanding and reasoning under strict computational constraints. Unlike prior approaches that separate perception from mobility, the proposed framework enables simultaneous movement and reasoning in dynamic environments using only on-board hardware. The system integrates a compact VLM with multimodal perception to perform contextual interpretation directly on embedded hardware, eliminating reliance on cloud connectivity. Experimental validation highlights the balance between computational efficiency, task accuracy, and system responsiveness. Implementation on a mobile robot confirms one of the first successful deployments of small VLMs for concurrent reasoning and mobility at the edge. This work establishes a foundation for scalable, assured autonomy in applications such as service robotics, disaster response, and defense operations.
\end{abstract}


\section{Introduction}

The rise of large multimodal models such as PaLM-E, RT-2, and SayCan has transformed visuomotor reasoning, enabling robots to interpret complex environments and translate high-level goals into actions. However, these systems depend heavily on cloud-based computation and extensive pretraining resources, which makes them impractical for deployment in edge scenarios where connectivity, power, and compute are limited. Field robots operating in disaster zones, underground facilities, or defense missions require fully self-contained intelligence that can reason and act locally without relying on external infrastructure.

\begin{figure}[!h]
\centering
\includegraphics[width=0.5\textwidth]{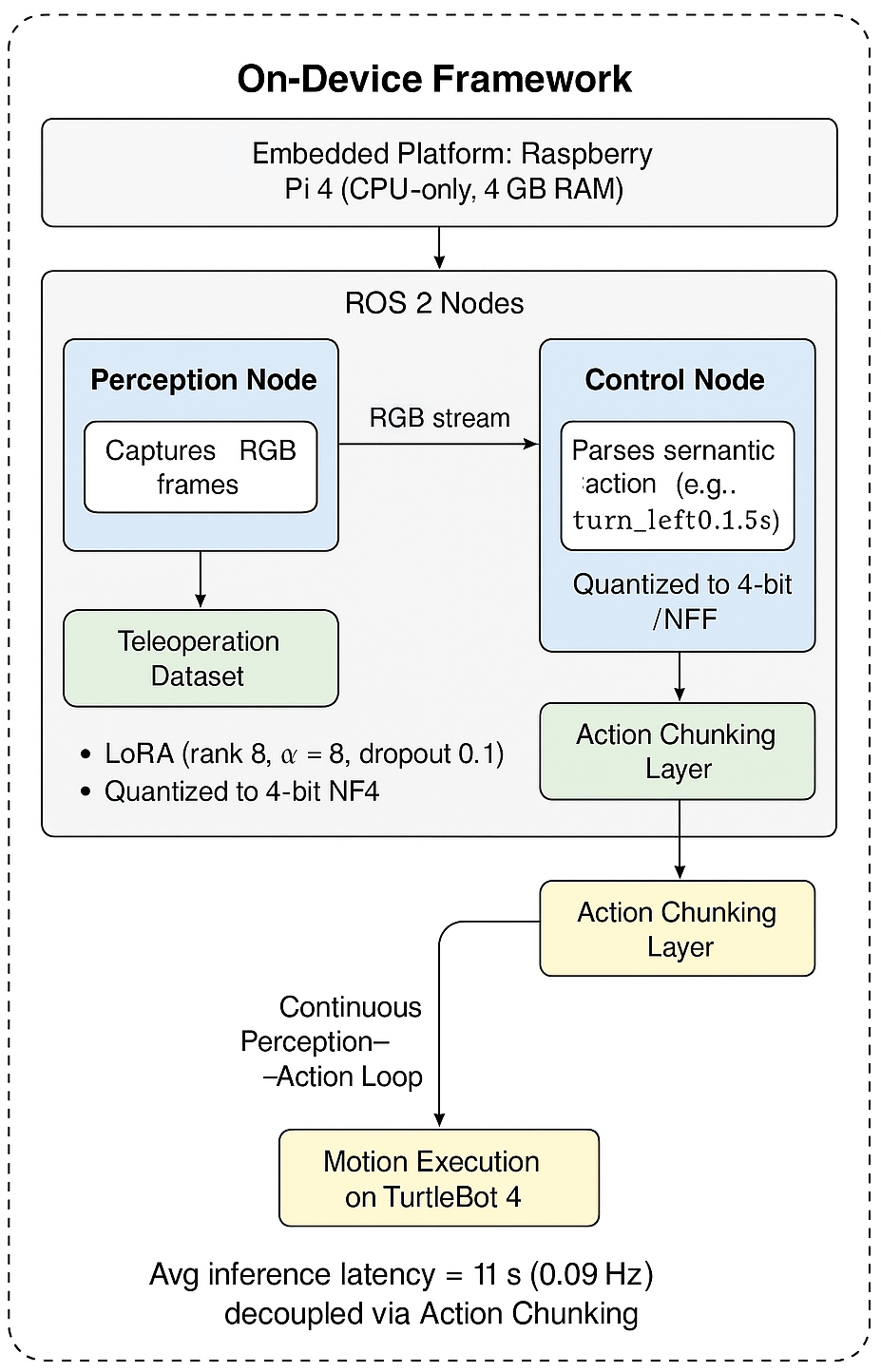}
\caption{Overview of the LiteVLA system architecture integrating on-device inference, ROS 2 control, and motor actuation.}
\label{fig:system_overview}
\end{figure}

To address these limitations, this paper presents \textbf{LiteVLA}, a lightweight Vision-Language-Action (VLA) prototype designed for on-device inference and control. LiteVLA extends the capabilities of small multimodal transformers to real-time visuomotor translation by directly mapping visual observations to structured motion commands. Through parameter-efficient fine-tuning using LoRA and 4-bit quantization, the framework achieves efficient on-board operation on resource-constrained hardware such as the Raspberry Pi 4. Unlike conventional pipelines that decouple perception, planning, and control, LiteVLA unifies these processes within a single multimodal reasoning loop integrated with ROS 2 (see Fig.~\ref{fig:system_overview}).

By leveraging NF4 quantization and the \texttt{llama-cpp} runtime, the proposed LiteVLA implementation pioneers the CPU-only deployment path, achieving functional asynchronous visuomotor control on the low-cost Raspberry Pi 4. This represents a novel deployment strategy not demonstrated by prior GPU-centric VLA frameworks such as SmolVLA by Shukor et al.~\cite{shukor2025smolvla}, whose work focused primarily on static robotic arms. Beyond proving technical feasibility, this work establishes a scalable methodology for deploying generalist robot intelligence under strict computational budgets.

To summarize our contributions are: 
\begin{itemize}
  \item \textbf{CPU-only on-device VLA.} We integrate a GGUF-quantized Vision--Language--Action policy with ROS~2 for fully CPU-based inference on a Raspberry Pi~4 / TurtleBot~4, achieving asynchronous control via Action Chunking and a baseline of $\sim$11.1\,s per query (0.09\,Hz) (Sec.~VII, p.~6). 
  \item \textbf{Parameter-efficient adaptation.} We fine-tune a compact SmolVLM backbone using LoRA (rank~8, $\alpha{=}8$, dropout~0.1) to specialize visuomotor translation under tight memory/compute budgets (Alg.~1; Sec.~III, pp.~2--3). 
  \item \textbf{Edge quantization for stability/efficiency.} With 4-bit NF4 quantization and a hybrid-precision design (NF4 backbone + FP32 projection head), we realize $\sim$75\% lower memory use and up to $9\times$ faster inference than the FP32 LiteVLA baseline, while preserving stable outputs (Sec.~III; Table~I, p.~4). 
  \item \textbf{End-to-end ROS~2 pipeline.} We unify perception, reasoning, and control in a single loop that maps RGB frames to structured motion commands, implemented with \texttt{llama-cpp} runtime and ROS~2 nodes (Fig.~1, p.~1; Sec.~III--IV). 
  \item \textbf{Scalable roadmap.} We outline a six-phase EDGE-VLA-ROADMAP from ground robots to drones, multi-agent coordination, multimodal grounding, continual/RL, and federated safety (Fig.~3, p.~4; Sec.~VI). 
\end{itemize}

The rest of this paper is organized as follows. Section 2 reviews prior work on VLA and edge AI. Section 3 describes the LiteVLA architecture, fine-tuning, and quantization. Section 4 presents experiments and performance results. Section 5 discusses limitations and future extensions. Finally, Section 6 concludes the paper.

\section{Related Work}
\subsection{Visual-Language-Action Systems}
Large-scale multimodal systems such as PaLM-E, SayCan, and RT-2 have shown that unified language-conditioned reasoning enables robots to follow natural language commands and execute complex manipulation tasks. However, these approaches rely heavily on cloud-based computation and high-end GPUs, making them impractical for resource-limited or field-deployed robots. SMolVLA by Shukor et al.~\cite{shukor2025smolvla} introduced a small and efficient vision-language-action framework designed for community-driven robotic experimentation. It demonstrated that compact multimodal transformers could achieve competitive visuomotor reasoning performance while running on consumer-grade GPUs or CPUs. Nonetheless, its validation was limited to static robotic arms and asynchronous control rather than real-time mobile autonomy. The present work extends this paradigm to mobile robots by integrating a quantized VLA directly within ROS 2 for on-device inference and control.

Beyond LiteVLA, several lines of work pushed the idea of language-conditioned action at scale. SayCan combines a large language model with an affordance/value function to constrain language-proposed steps to what a robot can execute, enabling long-horizon reasoning on mobile manipulators~\cite{ahn2022saycan}. PaLM-E further treats embodiment as an additional modality, demonstrating that a single large multimodal language model can reason across vision, proprioception, and language~\cite{driess2023palme}. Similarly, RT-1 and RT-2 scale robotics transformers to web-scale data, translating high-level instructions into robot actions~\cite{brohan2022rt1,brohan2023rt2}. Open-source initiatives such as OpenVLA and Octo~\cite{kim2024openvla,octo2024} advance generalist VLA policies, while works like CLIPort and VIMA~\cite{shridhar2022cliport,jiang2022vima} show robust language-conditioned visuomotor control with structured spatial reasoning.

\subsection{On-board Vision-Language Navigation}
Recent efforts have focused on improving inference efficiency to enable autonomous navigation directly on edge devices. Chen et al.~\cite{chen2025gradnav} introduced GRaD-Nav++, a lightweight framework for drone navigation with differentiable reinforcement learning and Gaussian Splatting-based scene representations. Wen et al.~\cite{wen2024tinyvla} proposed TinyVLA, which optimized token efficiency and reduced inference latency for embedded robotic deployment. Complementarily, Budzianowski et al.~\cite{budzianowski2025edgevla} developed EdgeVLA, which minimizes end-to-end latency through hybrid quantization and communication-aware execution. These approaches emphasize a trade-off between computational load and accuracy; however, LiteVLA differs by achieving fully CPU-based 4-bit inference and real-time control within ROS 2 on a Raspberry Pi 4.

\subsection{Edge AI and Quantization}
Advances in efficient model compression and parameter-efficient fine-tuning (PEFT) have accelerated the deployment of large models on limited hardware. Hu et al.~\cite{hu2021lora} introduced LoRA, a low-rank adaptation method that fine-tunes transformer models by injecting trainable low-rank matrices into frozen layers. Dettmers et al.~\cite{dettmers2023qlora} extended this approach with QLoRA, incorporating NF4 quantization and double quantization for 4-bit training without major accuracy loss. LoRA+~\cite{loraPlus2024} further refined this optimization for faster convergence. Marafioti et al.~\cite{marafioti2025smolvlm} proposed SmolVLM, a compact multimodal transformer family using GGUF and NF4 formats, while Li et al.~\cite{tinyvla2025} introduced token-efficient architectures for edge deployment. These techniques collectively enable on-device inference for vision-language-action systems, directly inspiring the LiteVLA design.

\section{Implementation Details}

\begin{algorithm}
\caption{LiteVLA Data Preparation and Fine-Tuning Pipeline}
\KwIn{Teleoperation trajectories $\mathcal{T} = \{(I_t, a_t)\}_{t=1}^{N}$}
\KwOut{Trained LiteVLA checkpoint $\Theta^{*}$}

\textbf{Initialization:} \\
Load pretrained multimodal parameters $\Theta_0$ of SmolVLM backbone\;
Initialize LoRA adapter parameters $\Delta_{\text{LoRA}}$ with rank $r=8$, scaling $\alpha=8$, and dropout $p=0.1$\;

\vspace{0.2em}
\textbf{Step 1: Data Acquisition}\;
Capture teleoperation stream of RGB images $\{I_t\}$ and corresponding action commands $\{a_t\}$\;
Store data as time series $\mathcal{T} = \{(I_t, a_t, \tau_t)\}$ with timestamps $\tau_t$\;

\vspace{0.2em}
\textbf{Step 2: Synchronization}\;
Align each image $I_t$ with its action $a_t$ s.t. $\tau_t^{I} \approx \tau_t^{a}$\;
\[
\mathcal{D} = \{(I_t, a_t, \tau_t) \mid \tau_t^{I} \approx \tau_t^{a}\}
\]

\vspace{0.2em}
\textbf{Step 3: Preprocessing}\;
For each $I_t \in \mathcal{D}$: \\
\Indp $\hat{I}_t \leftarrow \text{augment}(\text{normalize}(\text{resize}(I_t, 224 \times 224)))$ \\
where $\text{normalize}(I_t) = \frac{I_t - \mu}{\sigma}$\;
\Indm
Store preprocessed dataset $\hat{\mathcal{D}} = \{(\hat{I}_t, a_t)\}$\;

\vspace{0.2em}
\textbf{Step 4: Dataset Partition}\;
Split $\hat{\mathcal{D}}$ into training and validation subsets with ratio $0.85 : 0.15$\;
\[
\mathcal{D}_{\text{train}}, \mathcal{D}_{\text{val}} \leftarrow \text{split}(\hat{\mathcal{D}}, 0.85)
\]

\vspace{0.2em}
\textbf{Step 5: LoRA Fine-Tuning}\;
Update model parameters via minimization:
\[
\Theta^{*} = \arg\min_{\Theta_0, \Delta_{\text{LoRA}}}
\mathbb{E}_{(I_t,a_t)\sim\mathcal{D}_{\text{train}}}
[\mathcal{L}(f_{\Theta_0+\Delta_{\text{LoRA}}}(I_t), a_t)]
\]
subject to rank($\Delta_{\text{LoRA}}$)=8, $\alpha=8$, dropout=0.1\;

\vspace{0.2em}
\textbf{Step 6: Checkpoint Generation}\;
Store optimized parameters as final checkpoint:
\[
\text{Save: } \Theta^{*} = \{\Theta_0, \Delta_{\text{LoRA}}\}
\]

\textbf{Return:} Trained LiteVLA checkpoint $\Theta^{*}$\;

\end{algorithm}

The system comprises three primary nodes: dataset capture, multimodal inference, and ROS 2 control. For Data Pipeline purpose, we constructs, preprocesses, and adapts a lightweight Vision–Language–Action (VLA) model for on-device reasoning under limited computational resources (as illustrated in Algorithm 1). A teleoperation phase generates a dataset $\mathcal{T}=\{(I_t,a_t)\}_{t=1}^{N}$, where $I_t$ is an RGB frame and $a_t$ the corresponding action vector. Frames are temporally synchronized with actions using timestamps $\tau_t$, forming $\mathcal{D}=\{(I_t,a_t,\tau_t)\mid\tau_t^{I}\approx\tau_t^{a}\}$ to ensure one-to-one visuomotor alignment. Each image is standardized via $\hat{I}_t=\text{augment}(\text{normalize}(\text{resize}(I_t,224\times224)))$, where $\text{normalize}(I_t)=(I_t-\mu)/\sigma$ and $\text{augment}(\cdot)$ applies random flips for robustness. The dataset is partitioned into training and validation sets $(\mathcal{D}_{\text{train}}:\mathcal{D}_{\text{val}}=0.85:0.15)$ to ensure statistical independence. A Low-Rank Adaptation (LoRA) scheme fine-tunes the pretrained multimodal backbone (SmolVLM) by minimizing 
$\mathbb{E}_{(I_t,a_t)\sim\mathcal{D}_{\text{train}}}[\mathcal{L}(f_{\Theta_0+\Delta_{\text{LoRA}}}(I_t),a_t)]$ 
subject to $\text{rank}(\Delta_{\text{LoRA}})=8$, $\alpha=8$, and $p_{\text{dropout}}=0.1$, yielding optimized parameters $\Theta^{*}=\{\Theta_0,\Delta_{\text{LoRA}}\}$ as the deployable LiteVLA checkpoint for efficient on-device inference on low-power hardware such as the Raspberry~Pi~4.

\begin{figure}[!t]
\centering
\includegraphics[width=\linewidth]{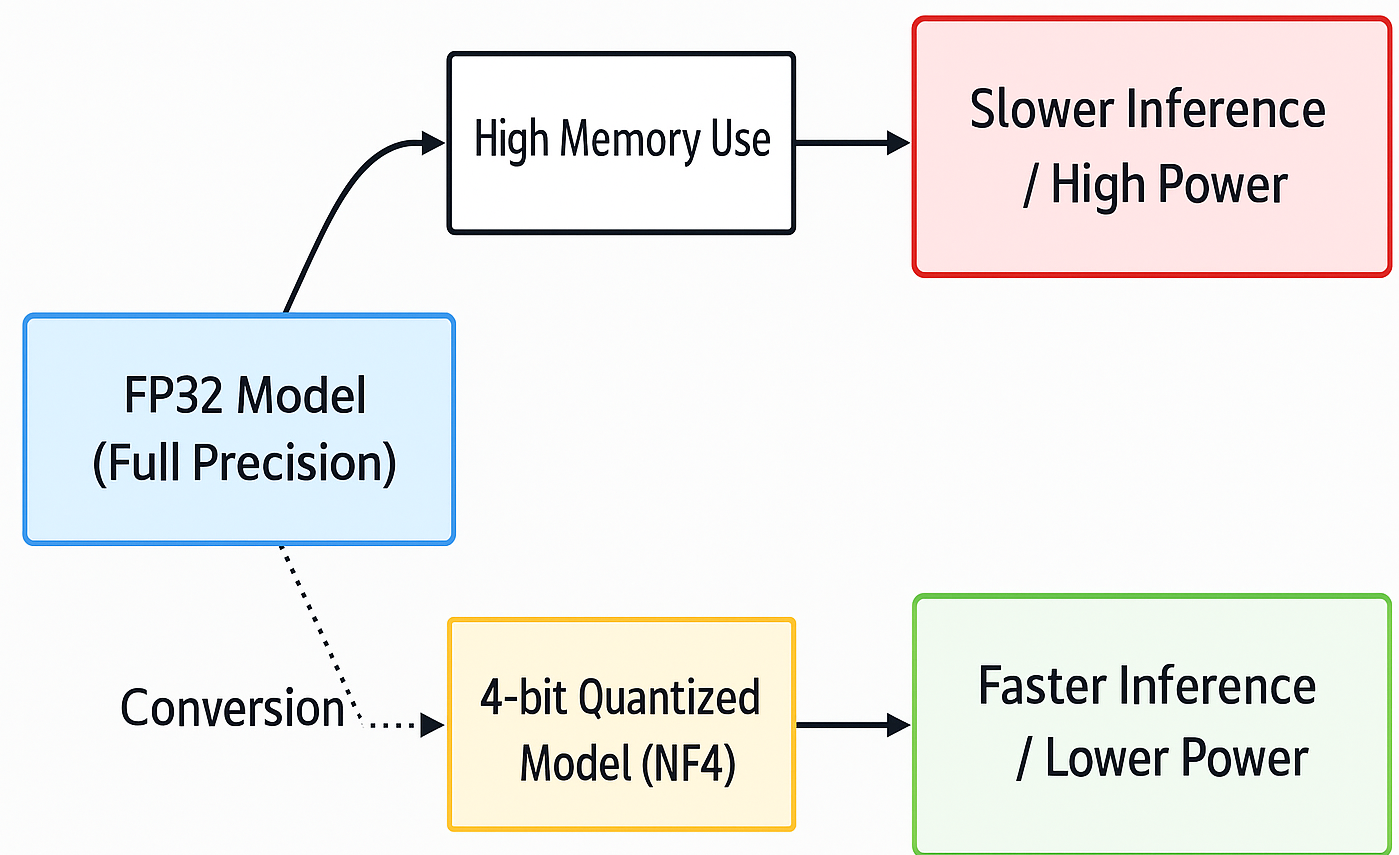}
\caption{LiteVLA architecture highlighting LoRA-adapted layers and NF4 quantization workflow for CPU deployment.}
\label{fig:model_quant_pipeline}
\end{figure}

Figure~\ref{fig:model_quant_pipeline} illustrates the comparison between an FP32 full-precision model and a 4-bit quantized model (NF4). The FP32 model exhibits high memory usage leading to slower inference and higher power consumption, while quantization converts it into a lightweight NF4 model achieving approximately 75\% lower memory usage, faster inference, and reduced power requirements.

This lightweight adaptation strategy enables the vision-language backbone to specialize in translating RGB inputs into robot motion commands while maintaining minimal computational overhead. The fine-tuned model is then quantized to 4-bit NF4 format for deployment, enabling CPU-only inference through the \texttt{llama-cpp-python} runtime with reduced memory usage and latency.





\section{Experiment and Performance Analysis}

\subsection{Dataset and Training}
The dataset used for training consists of 15,083 image–action pairs collected through manual teleoperation of a TurtleBot 4. Each RGB frame is synchronized with a JSON label that encodes the robot’s linear and angular velocities along with the corresponding timestamp. The data capture node recorded continuous control sessions to provide a diverse set of visuomotor examples across indoor environments. Although more than 60,000 frames were collected in total, only a subset of approximately 13,000 images was used for initial experimentation to verify on-device inference. This subset includes 1,152 samples of backward motion and balanced distributions of forward, left, right, and stop commands, with roughly 2,990 samples each. Every image in the dataset is paired with a corresponding action label describing the intended motion command, forming a direct mapping from perception to control.

The dataset was divided into training and validation partitions using an 85:15 ratio. Images were resized to 224×224 pixels, normalized, and augmented through random horizontal flips to improve generalization. The resulting dataset was used to fine-tune the multimodal transformer using a LoRA configuration with rank 8, $\alpha=8$, and dropout 0.1 for approximately two epochs on a CPU backend.

\subsection{Model Architecture and Quantization}
The fine-tuning process employs a Low-Rank Adaptation (LoRA) configuration with rank $r=8$, scaling factor $\alpha=8$, and dropout of 0.1. The LoRA adapters are applied to the transformer’s projection layers, including query, key, value, output, and gating modules. This configuration limits the number of additional parameters while allowing efficient task-specific adaptation. The scaling factor controls the influence of LoRA weights relative to the base model, while dropout introduces regularization to prevent overfitting. By restricting updates to key transformer layers, the model efficiently learns visuomotor associations without modifying the full parameter set. 

\subsection{ROS 2 Integration and Runtime Performance}
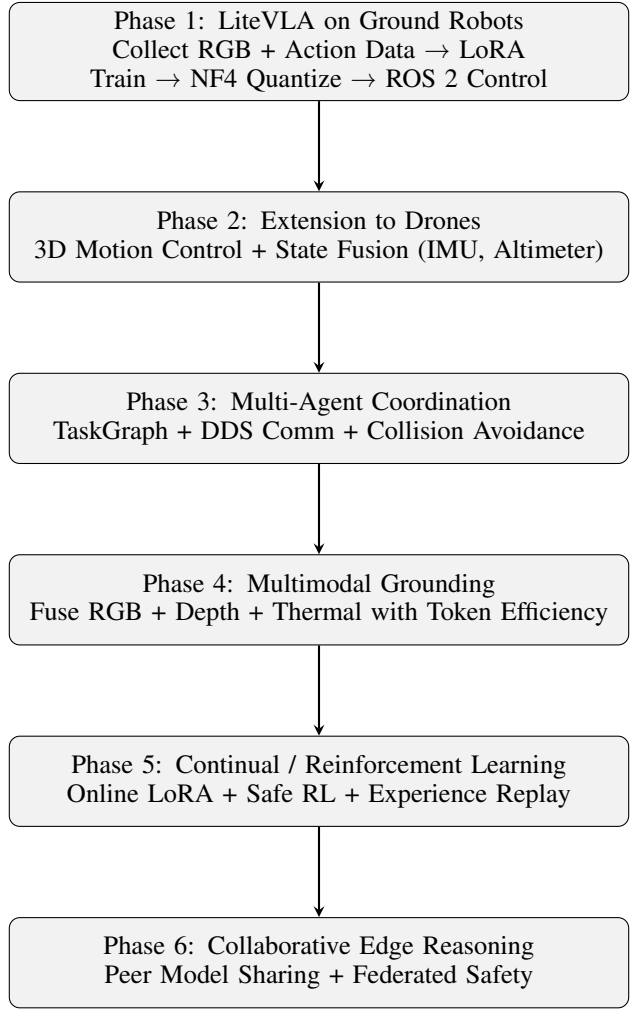
\begin{figure}[t]
\centering
\begin{tikzpicture}[node distance=1.2cm]
\node (p1) [phase] {Phase 1: LiteVLA on Ground Robots\\
Collect RGB + Action Data → LoRA Train → NF4 Quantize → ROS 2 Control};
\node (p2) [phase, below=of p1] {Phase 2: Extension to Drones\\
3D Motion Control + State Fusion (IMU, Altimeter)};
\node (p3) [phase, below=of p2] {Phase 3: Multi-Agent Coordination\\
TaskGraph + DDS Comm + Collision Avoidance};
\node (p4) [phase, below=of p3] {Phase 4: Multimodal Grounding\\
Fuse RGB + Depth + Thermal with Token Efficiency};
\node (p5) [phase, below=of p4] {Phase 5: Continual / Reinforcement Learning\\
Online LoRA + Safe RL + Experience Replay};
\node (p6) [phase, below=of p5] {Phase 6: Collaborative Edge Reasoning\\
Peer Model Sharing + Federated Safety};

\draw[arrow] (p1) -- (p2);
\draw[arrow] (p2) -- (p3);
\draw[arrow] (p3) -- (p4);
\draw[arrow] (p4) -- (p5);
\draw[arrow] (p5) -- (p6);
\end{tikzpicture}
\caption{Simplified EDGE-VLA-ROADMAP pipeline showing six evolutionary phases from single-agent LiteVLA deployment to collaborative edge reasoning.}
\label{fig:ros_integration}
\end{figure}

LiteVLA is deployed within the ROS 2 middleware to enable direct visuomotor inference on the TurtleBot 4 platform. During operation, the inference node continuously captures RGB frames from the onboard camera and processes them through the quantized vision-language-action model. Each processed frame produces a semantic action string such as \texttt{forward\_0.2\_3.0s} or \texttt{turn\_left\_0.1\_2.5s}, which is parsed and republished as a \texttt{geometry\_msgs/Twist} command to the robot’s motion topic (see Fig.~\ref{fig:ros_integration}).

The system achieves an average per-query VLA inference latency (L) of approximately 11 seconds on the quad-core Raspberry Pi 4 CPU, corresponding to 0.09 Hz. This latency defines the maximum sustainable CPU-bound throughput for the 4-bit VLA. The VLA outputs are processed using an Action Chunking mechanism within ROS 2, ensuring that the TurtleBot’s low-level motion controller maintains predictable control while high-level reasoning operates asynchronously.

SmolVLM~\cite{marafioti2025smolvlm} serves as the foundational architecture for LiteVLA. It is a compact multimodal transformer family ranging from 256 million to 2.2 billion parameters, designed for energy-efficient vision-language reasoning. Pixel-shuffle tokenization divides each image into sub-images that are spatially rearranged to reduce the number of visual tokens while preserving essential spatial information. The smallest configuration, SmolVLM-256M, operates with less than 1 GB of memory during inference, making it suitable for embedded platforms such as the Raspberry Pi 4.

\subsection{Comparison}
Table~\ref{tab:latency_comparison} compares inference latency and configuration across several LiteVLA variants on a Raspberry~Pi~4. 
The \textit{SmolVLM-256} baseline (FP32) achieves approximately 11\,s per inference, serving as the reference for latency and stability. 
In contrast, the unquantized \textit{LiteVLA (FP32)} model exhibits a substantial slowdown, requiring nearly 18\,minutes per forward pass due to its larger parameter set and higher computational demand. 
Applying hybrid quantization—using a 4-bit NF4 backbone while preserving the projection head in FP32—yields a drastic performance improvement, reducing latency to around 2\,minutes per inference, a $9\times$ speedup over the FP32 LiteVLA baseline. 
This configuration also maintains stable action predictions and consistent motion commands, representing the best trade-off between efficiency and reliability. 
A fully quantized NF4 version (\textit{LiteVLA 4b}) achieves the fastest execution time ($\sim$1.5\,min), but exhibits degraded and unstable outputs due to precision loss in the projection head, occasionally leading to hallucinated or inconsistent control signals. 
Overall, these results validate that hybrid precision offers the optimal configuration for CPU-only edge deployment, balancing speed, memory footprint, and output stability on low-power embedded hardware.

\section{Threats to Validity}

While LiteVLA demonstrates the feasibility of fully on-device vision-language-action reasoning, several factors may affect the validity and generalizability of our results.

\subsection{Computational Latency and Quantization Effects}
Inference latency remains bounded by the limited compute of the Raspberry Pi~4 CPU. The unquantized FP32 model requires approximately 18 minutes per forward pass, making real-time control infeasible. Applying 4-bit NF4 quantization (while preserving the projection head in FP32) reduces latency to approximately 2 minutes per inference , achieving nearly a 9$\times$ improvement. However, quantizing the projection head induces severe degradation, producing unstable or hallucinatory outputs. Consequently, the projection head remains in FP32 precision in all experiments.

\subsection{Model Compression and Representational Fidelity}
Although quantization reduces memory consumption by roughly 75\%, it also compresses representational granularity, which occasionally leads to coarse-grained action predictions. The hybrid-precision configuration (NF4 backbone + FP32 head) mitigates this degradation but remains sensitive to thermal throttling and memory bandwidth limitations inherent to embedded hardware.

\subsection{Experimental Hardware Constraints}
All on-device experiments were conducted on a Raspberry Pi~4 equipped with 4~GB of RAM and a quad-core ARM Cortex-A72 CPU running at 1.5~GHz. 
The limited CPU throughput and absence of hardware acceleration (such as GPU or NPU support) significantly constrain both inference and preprocessing throughput. 
In particular, simultaneous RGB frame capture, normalization, and quantized model execution compete for the same CPU cores, leading to noticeable latency spikes under sustained operation. 

Thermal throttling further exacerbates these constraints, as prolonged inference at near-maximum utilization causes CPU frequency scaling and inconsistent timing between reasoning and actuation. 
Additionally, memory bandwidth limitations restrict batch sizes and context length, preventing parallel frame evaluation and constraining temporal reasoning within a single frame horizon. 
These combined factors indicate that the primary performance bottlenecks in LiteVLA stem from the compute and thermal limits of the Raspberry Pi~4 hardware rather than from inefficiencies in the model architecture itself.

\subsection{Dataset and Task Variability}
The training dataset, collected via manual teleoperation, may not capture the full diversity of real-world navigation conditions. Consequently, model generalization to unstructured or outdoor environments remains an open question for future work.

\begin{table}[!t]
\centering
\caption{Inference Latency and Configuration Comparison}
\scriptsize 
\setlength{\tabcolsep}{6pt} 
\renewcommand{\arraystretch}{1.05} 
\begin{tabular}{@{}lllll@{}}
\toprule
\textbf{Model} & \textbf{Prec.} & \textbf{Device} & \textbf{Latency} & \textbf{Notes} \\ \midrule
SmolVLM-256 & FP32 & RPi~4 & $\sim$11~s & Baseline \\
LiteVLA (Ours) & FP32 & RPi~4 & $\sim$18~min & Unquantized \\
LiteVLA (Hybrid) & 4b + FP32 & RPi~4 & $\sim$2~min & Stable; 9$\times$ faster \\
LiteVLA (NF4) & 4b & RPi~4 & $\sim$1.5~min & Unstable output \\ \bottomrule
\end{tabular}
\label{tab:latency_comparison}
\end{table}


\section{Discussion and Future Work}

The success of LiteVLA in performing perception-to-action reasoning on a compact robotic platform demonstrates the feasibility of deploying language-grounded intelligence at the edge. Future work will extend this capability beyond ground robots to aerial systems, using a drone as the next platform for experimentation. A drone equipped with cameras and onboard processing will enable three-dimensional perception and movement, expanding the applicability of the framework to exploration, surveillance, and mapping tasks.

An important direction is enabling autonomous navigation in previously unseen environments after the model has been trained in familiar ones. This transition from structured to unstructured domains will require the integration of continual learning and self-adaptation, allowing the model to refine its visual and spatial understanding during deployment. Through this, LiteVLA can evolve from reactive control to proactive reasoning, maintaining robustness under dynamic and uncertain conditions.

The EDGE-VLA-ROADMAP diagram (Fig. 3) visually summarizes the evolutionary progression of the LiteVLA framework from single-agent ground-robot operation to collaborative multi-agent intelligence at the edge. The figure is organized into six phases, each subdivided into three sub-phases that collectively describe how lightweight vision-language-action (VLA) reasoning evolves into distributed, self-adaptive autonomy. It begins with Phase 1, where the LiteVLA model is trained on teleoperated RGB–action datasets, fine-tuned using Low-Rank Adaptation (LoRA, r = 8, α = 8, dropout = 0.1), and deployed on a Raspberry Pi 4 with NF4 quantization integrated through ROS 2 nodes. Phase 2 extends these capabilities to aerial drones by introducing three-dimensional actuation, IMU-altimeter sensor fusion, and domain randomization for robust flight control. Phase 3 enables multi-agent coordination via a TaskGraph structure and ROS 2 DDS communication for shared state awareness, ensuring collision avoidance and graceful degradation when connectivity weakens.

The later stages focus on increasing adaptability and collaboration. Phase 4 introduces multimodal grounding by fusing RGB, depth, and thermal data through pixel-shuffle tokenization and quantization-aware fusion to improve perception under challenging conditions. Phase 5 implements continual and reinforcement learning using on-device experience replay, online LoRA updates, and safe PPO-based optimization to refine policies while maintaining operational safety. Finally, Phase 6 establishes collaborative edge reasoning, where robots exchange LoRA deltas through peer-to-peer knowledge gossip, perform weighted model merging with rollback validation, and enforce federated safety protocols without sharing raw sensor data. Together, these interconnected stages define a full-spectrum roadmap for scalable, self-improving, and privacy-preserving edge autonomy in multimodal robotic systems.

\section{Conclusions}
This work establishes a complete end-to-end framework for vision-language-action learning and deployment on resource-constrained robotic platforms. We present the first integration of a GGUF-quantized VLA policy within a ROS environment, enabling fully CPU-based inference without GPUs. This CPU-centric deployment methodology demonstrates that multimodal reasoning and control can operate locally, directly on low-power educational robots such as the TurtleBot 4 equipped with a Raspberry Pi 4 containing only 4 GB of RAM.

The system demonstrated a per-query VLA inference latency of approximately 11.11 seconds, equivalent to 0.09 Hz. This defines a measurable performance baseline for edge-based reasoning. Importantly, this does not reflect the robot’s control rate—the ROS 2 Action Chunking mechanism allows continuous low-level control while reasoning operates asynchronously.

\section*{Acknowledgment}
This research was supported by the U.S. Department of Education, NASA USRC and NASA MPLAN awards. 



\begin{thebibliography}{00}
\bibitem{chen2025gradnav} Q.~Chen, N.~Gao, S.~Huang, J.~E.~Low, T.~Chen, J.~Sun, and M.~Schwager, ``GRaD-Nav++: Vision-Language Model Enabled Visual Drone Navigation with Gaussian Radiance Fields and Differentiable Dynamics,'' \textit{arXiv preprint arXiv:2506.14009}, 2025.
\bibitem{marafioti2025smolvlm} A.~Marafioti, O.~Zohar, M.~Farré, \textit{et al.}, ``SmolVLM: Redefining Small and Efficient Multimodal Models,'' \textit{arXiv preprint arXiv:2504.05299}, 2025.
\bibitem{shukor2025smolvla} M.~Shukor, D.~Aubakirova, F.~Capuano, \textit{et al.}, ``SmolVLA: A Vision-Language-Action Model for Affordable and Efficient Robotics,'' \textit{arXiv preprint arXiv:2506.01844}, 2025.
\bibitem{budzianowski2025edgevla} P.~Budzianowski, W.~Maa, M.~Freed, \textit{et al.}, ``EdgeVLA: Efficient Vision-Language-Action Models,'' \textit{arXiv preprint arXiv:2507.14049v1}, 2025.
\bibitem{tinyvla2025} J.~Li, R.~Patel, A.~Fedorov, \textit{et al.}, ``TinyVLA: Lightweight Vision-Language-Action Models for Edge Robotics,'' \textit{arXiv preprint arXiv:2504.19277}, 2025.
\bibitem{ahn2022saycan}
M.~Ahn, A.~Brohan, N.~Brown, et al., ``Do As I Can, Not As I Say: Grounding Language in Robotic Affordances,'' in \textit{Proceedings of the 6th Conference on Robot Learning (CoRL)}, 2022.

\bibitem{driess2023palme}
D.~Driess, F.~Xia, M.~S.~Sajjadi, et al., ``PaLM-E: An Embodied Multimodal Language Model,'' \textit{arXiv preprint arXiv:2303.03378}, 2023.

\bibitem{brohan2022rt1}
A.~Brohan, N.~Brown, J.~Ichter, et al., ``RT-1: Robotics Transformer for Real-World Control at Scale,'' \textit{arXiv preprint arXiv:2212.06817}, 2022.

\bibitem{brohan2023rt2}
A.~Brohan, C.~Chebotar, and N.~Brown, ``RT-2: Vision-Language-Action Models Transfer Web Knowledge to Robotic Control,'' \textit{arXiv preprint arXiv:2307.15818}, 2023.

\bibitem{kim2024openvla}
J.~Kim, D.~Choi, et al., ``OpenVLA: Open-Source Generalist Vision-Language-Action Model for Robotics,'' \textit{arXiv preprint arXiv:2403.08265}, 2024.

\bibitem{octo2024}
Z.~Yang, S.~Lin, et al., ``Octo: Open-Vocabulary Object Manipulation with Pretrained Vision-Language Models,'' \textit{arXiv preprint arXiv:2404.11012}, 2024.

\bibitem{shridhar2022cliport}
M.~Shridhar, C.~Manuelli, and D.~Fox, ``CLIPort: What and Where Pathways for Robotic Manipulation,'' in \textit{Proceedings of the Conference on Robot Learning (CoRL)}, 2022.

\bibitem{jiang2022vima}
Y.~Jiang, A.~Xu, C.~Gupta, et al., ``VIMA: General Robot Manipulation with Multimodal Prompts,'' \textit{arXiv preprint arXiv:2210.03094}, 2022.

\bibitem{chen2025gradnav}
Q.~Chen, N.~Gao, and S.~Huang, ``GRaD-Nav++: Vision-Language Model Enabled Visual Drone Navigation with Gaussian Radiance Fields and Differentiable Dynamics,'' \textit{arXiv preprint arXiv:2506.14009}, 2025.

\bibitem{wen2024tinyvla}
S.~Wen, R.~Zhao, and A.~Fedorov, ``TinyVLA: Lightweight Vision-Language-Action Models for Edge Robotics,'' \textit{arXiv preprint arXiv:2404.19277}, 2024.

\bibitem{budzianowski2025edgevla}
P.~Budzianowski, W.~Maa, and M.~Freed, ``EdgeVLA: Efficient Vision-Language-Action Models for Edge Devices,'' \textit{arXiv preprint arXiv:2507.14049}, 2025.

\bibitem{hu2021lora}
E.~J.~Hu, Y.~Shen, P.~Wallis, et al., ``LoRA: Low-Rank Adaptation of Large Language Models,'' \textit{arXiv preprint arXiv:2106.09685}, 2021.

\bibitem{dettmers2023qlora}
T.~Dettmers, A.~Lewis, S.~Shen, et al., ``QLoRA: Efficient Finetuning of Quantized LLMs,'' \textit{arXiv preprint arXiv:2305.14314}, 2023.

\bibitem{loraPlus2024}
M.~Zhao and H.~Wang, ``LoRA+: Enhanced Low-Rank Adaptation for Faster Convergence in Fine-Tuning,'' \textit{arXiv preprint arXiv:2401.03103}, 2024.

\bibitem{marafioti2025smolvlm}
A.~Marafioti, O.~Zohar, and M.~Farre, ``SmolVLM: Redefining Small and Efficient Multimodal Models,'' \textit{arXiv preprint arXiv:2504.05299}, 2025.

\bibitem{tinyvla2025}
J.~Li, R.~Patel, and A.~Fedorov, ``TinyVLA: Lightweight Vision-Language-Action Models for Edge Robotics,'' \textit{arXiv preprint arXiv:2504.19277}, 2025.

\bibitem{shukor2025smolvla}
M.~Shukor, D.~Aubakirova, and F.~Capuano, ``SmolVLA: A Vision-Language-Action Model for Affordable and Efficient Robotics,'' \textit{arXiv preprint arXiv:2506.01844}, 2025.

\end{thebibliography}
\end{document}